\documentclass{article}

\usepackage{PRIMEarxiv}
\usepackage{multirow}
\usepackage[utf8]{inputenc} 
\usepackage[T1]{fontenc}    
\usepackage{hyperref}       
\usepackage{url}            
\usepackage{booktabs}       
\usepackage{amsfonts}       
\usepackage{nicefrac}       
\usepackage{microtype}      
\usepackage{lipsum}
\usepackage{fancyhdr}       
\usepackage{graphicx}       
\graphicspath{{media/}}     

\title{Development and evaluation of a deep learning algorithm for German word recognition from lip movements
\thanks{\textit{English version of}: 
\textbf{Pham, D.N., Rahne, T. Entwicklung und Evaluation eines Deep-Learning-Algorithmus für die Worterkennung aus Lippenbewegungen für die deutsche Sprache. HNO 70, 456–465 (2022). https://doi.org/10.1007/s00106-021-01143-9.}} 
}

\author{
  Dinh Nam Pham\\
  Technische Universität Berlin \\
  Berlin, Germany \\
  \texttt{dinh-nam.pham@campus.tu-berlin.de} \\
   \And
  Torsten Rahne \\
  Department of Otorhinolaryngology\\
  University Medicine Halle (Saale) \\
  Martin Luther University Halle-Wittenberg \\
  Halle (Saale), Germany \\
  \texttt{torsten.rahne@uk-halle.de} \\
}

\begin{document}
\maketitle

\begin{abstract}
When reading lips, many people benefit from additional visual information from the lip movements of the speaker, which is, however, very error prone. Algorithms for lip reading with artificial intelligence based on artificial neural networks significantly improve word recognition but are not available for the German language. A total of 1806 videoclips with only one German-speaking person each were selected, split into word segments, and assigned to word classes using speech-recognition software. In 38,391 video segments with 32 speakers, 18 polysyllabic, visually distinguishable words were used to train and validate a neural network. The 3D Convolutional Neural Network and Gated Recurrent Units models and a combination of both models (GRUConv) were compared, as were different image sections and color spaces of the videos. The accuracy was determined in 5000 training epochs. Comparison of the color spaces did not reveal any relevant different correct classification rates in the range from 69\% to 72\% . With a cut to the lips, a significantly higher accuracy of 70\%  was achieved than when cut to the entire speaker’s face (34\% ). With the GRUConv model, the maximum accuracies were 87\%  with known speakers and 63\%  in the validation with unknown speakers. The neural network for lip reading, which was first developed for the German language, shows a very high level of accuracy, comparable to English-language algorithms. It works with unknown speakers as well and can be generalized with more word classes.
\end{abstract}

\keywords{Visual Speech Recognition \and Lipreading \and Deep Learning \and Feature Extraction \and Neural
 Networks}

\section*{Preface}
This work is an English translation of the original German-language journal article published in January 2022 (DOI:10.1007/S00106-021-01143-9), intended to make its content accessible to the international research community. To the best of our knowledge, this represented the first ever peer-reviewed publication on automatic lip reading for the German language. We kindly request that readers cite the original publication rather than this translated version.

\section{Background and Research Question}
A large number of people in the German-speaking area suffer from hearing loss \cite{b9}. The more severe and prolonged the hearing loss, the more they benefit from additional cues derived from the speaker’s lip movements \cite{b1,b26}. Even those with normal hearing can rely on this to improve comprehension in challenging auditory environments, particularly under noisy conditions \cite{b24,b36}. However, this so-called lip-reading remains error-prone. In German, only about 15\% of speech sounds can be visually distinguished from lip patterns alone. When contextual clues are incorporated, accuracy rises to just 50\% \cite{b18}.

In recent years, artificial intelligence (AI) algorithms based on artificial neural networks have shown promise in enhancing lip-reading \cite{b14}. For English, AI-driven approaches like LipNet \cite{b2} and Google DeepMind \cite{b27} have been employed. Yet, no such applications currently exist for the German language.

Tasks like lip-reading present significant challenges, as no known algorithms currently solve them completely or optimally. Even humans lack the ability to flawlessly translate lip movements into speech in every instance. However, lip-reading demonstrates substantial learning effects through practice \cite{b15,b32}. To achieve the highest possible accuracy in lip-reading with computers, researchers have turned to deep learning methods based on multi-layered artificial neural networks (ANNs). These systems adjust the weights between neurons using training data, ultimately developing the ability to generalize—applying learned knowledge to new examples with reduced error rates. Neural networks function as universal approximators; theoretically, given sufficient neurons, they can model n-dimensional functions \cite{b12}. Assuming that lip-reading involves detectable patterns or correlations, the success of ANNs in learning this task becomes explicable.

As a core component of most deep learning models, the Multilayer Perceptron (MLP) serves as a universal function approximator \cite{bCybenko1989}, enabling tasks such as regression analysis and classification. Neurons—the smallest computational units in such networks—have n inputs, which may originate from other perceptrons. Each input is assigned a weight $w_i$. The perceptron’s output is computed by applying an activation function to the weighted sum of its inputs, augmented by a weighted bias term.
To enhance predictive power, multiple perceptrons are combined into a multilayer perceptron network. Each perceptron in a given layer receives the outputs of the preceding layer as its inputs. To approximate nonlinear complex relationships, neural networks require nonlinear activation functions. These enable models to capture more sophisticated input-output mappings which is an essential capability for tasks such as lip-reading.. In this work, the most widely used Rectified Linear Unit (ReLU) function \cite{b34} (for hidden layers) and the Softmax function \cite{b3} (for output layers) are employed. The ReLU function outputs its argument if positive; otherwise, it returns zero. Softmax—a normalized exponential function—computes a probability distribution over a vector of real-valued components, facilitating multiclass classification at the output layer. Here, each component represents a class, allowing the model to identify the most probable class.
Collectively, an MLP with sufficient neurons, layers, and nonlinear activation functions can approximate arbitrary n-dimensional functions.

Besides MLP, another class of neural networks includes 2D convolutional layers, which consist of a predefined number of filters (kernels) \cite{b21}. These filters are matrices with real-valued elements and fixed dimensions. Both kernel size and the number of filters are key parameters to consider when designing a Convolutional Neural Network (CNN).
In a convolutional layer, filters slide incrementally across the input matrix. The kernel moves row-wise from left to right with a predetermined step size (stride). At each step, the elements of the input matrix are multiplied by their corresponding weights in the kernel, and the resulting products are summed. This sum becomes a single element in the output feature map. Updating the filter weights enables the network to learn pattern recognition and feature extraction. The depth of the resulting feature map is determined by the number of kernels.
While 2D Convolutional Neural Networks (CNNs or ConvNets) were originally developed for computer vision tasks such as handwriting recognition \cite{b19}, 3D CNNs achieve higher accuracy with 3D image data \cite{b20}. Videos can be represented as four-dimensional arrays with dimensions for channels, depth, height, and width, making them suitable for 3D CNNs. Unlike MLPs, which only process vector inputs, CNNs can handle matrix inputs—including images. By employing multiple kernels and successive layers, CNNs can detect increasingly complex patterns. In the case of videos, 3D CNNs can extract not only spatial but possibly also temporal information.

Unlike the aforementioned feedforward layers \cite{b33}, where outputs are always passed to the next layer, recurrent neural networks (RNNs) with feedback loops are particularly well-suited for processing sequential data \cite{b7,b12}. Among these, Gated Recurrent Units (GRUs) \cite{b6} efficiently capture temporal information in videos with relatively low computational overhead.
A GRU employs two key mechanisms: the reset gate, implemented as a vector, determines which past information should be erased (or "forgotten"), while the update gate controls which information from previous hidden states should be retained. Through exposure to training examples, the network gradually learns to extract temporal patterns to optimize its predictions.
However, unlike CNNs, both GRUs and MLPs can only process vector inputs. When passing outputs from CNNs—which are three- or four-dimensional arrays—to these layers, the data must first be flattened into one-dimensional vectors.

Training robust automated lip-reading algorithms requires large-scale datasets, which currently do not exist for the German language. A model's performance critically depends on the size, quality, and relevance of its training data \cite{b14}. Current speech recognition accuracy from lip movements varies significantly across studies, with results being highly dependent on both the algorithm architecture and dataset characteristics.
For English, Chung et al. \cite{b7} achieved 47\% sentence-level accuracy on a 10,000-item dataset using a CNN-RNN hybrid architecture. Yang et al. \cite{b35} subsequently created an even larger dataset for Chinese. As Hao's \cite{b14} overview shows, reported word recognition rates for English range from 76\% \cite{b7} to 98\% \cite{b31}. However, no comparable algorithms or large-scale datasets currently exist for German.

The goal of this work is to develop automated word-level lip-reading for German speakers in the form of visual speech recognition by designing and implementing a suitable neural network. For this purpose, a dataset will be created to serve for training and evaluating the deep learning models. The accuracy will be examined for both speakers that were included in the training as well as unseen speakers. To achieve the highest possible accuracy, different network architectures and preprocessing methods for the input data will be compared. A processing pipeline will be developed that handles input data from speakers without audio and classifies it through the model.

\section{Materials and Methods}

\subsection{Dataset}

Initially, 1,806 videos were selected from the video platform YouTube (YouTube LLC, San Bruno, USA), each featuring only one clearly identifiable German-speaking person, and saved locally for further processing. Videos specifically intended for learning German were preferentially selected. Using the open-source speech recognition software Vosk API \cite{bvosk}, the spoken words were identified and their corresponding timestamps determined.
For the dataset, the following 18 multisyllabic, visually distinguishable words were used as word classes: comments - questions - exam - Germany - can - speak - really - actually - know - naturally - video - means - example - write - people - simple - important - words.

Thus, the complete dataset consists of 38,391 videos featuring 32 speakers, with each video showing one speaker articulating one of the 18 words. This dataset was divided into three non-overlapping subsets (A, B, C). Each subset contains videos from different speakers. Dataset A was further split into training and validation sets in an 8:2 ratio. Dataset B was divided into training, validation, and test sets in an 8:1:1 ratio (Table \ref{Tab:dataset}).

For both datasets, the class distribution was preserved during splitting, ensuring that the class with the most or fewest videos remained the same in both training and validation sets. This prevents a majority of videos from one class ending up in the validation set while leaving insufficient examples of that class in the training set for learning. Each video in the validation and test sets was manually reviewed, and videos containing incorrect words were removed.

\begin{table}[]
\centering
\begin{tabular}{|ll|l|l|l|}
\hline
\multicolumn{2}{|l|}{Dataset}                         & Original Videos       & Speakers            & Videos in the Dataset \\ \hline
\multicolumn{1}{|l|}{\multirow{3}{*}{A}} & Total      & \multirow{3}{*}{250}  & \multirow{3}{*}{6}  & 3727                  \\ \cline{2-2} \cline{5-5} 
\multicolumn{1}{|l|}{}                   & Training   &                       &                     & 2973                  \\ \cline{2-2} \cline{5-5} 
\multicolumn{1}{|l|}{}                   & Valdiation &                       &                     & 754                   \\ \hline
\multicolumn{1}{|l|}{\multirow{3}{*}{B}} & Total      & \multirow{3}{*}{1400} & \multirow{3}{*}{22} & 30.684                \\ \cline{2-2} \cline{5-5} 
\multicolumn{1}{|l|}{}                   & Training   &                       &                     & 24.553                \\ \cline{2-2} \cline{5-5} 
\multicolumn{1}{|l|}{}                   & Validation &                       &                     & 3060                  \\ \hline
\multicolumn{1}{|l|}{\multirow{2}{*}{C}} & Total      & \multirow{2}{*}{156}  & \multirow{2}{*}{4}  & 3950                  \\ \cline{2-2} \cline{5-5} 
\multicolumn{1}{|l|}{}                   & Test       &                       &                     & 3950                  \\ \hline
\end{tabular}
\caption{Structure and split of the datasets.}
\label{Tab:dataset}
\end{table}

\subsection{Video Processing}

\begin{figure}[h]
\centering
\includegraphics[scale=0.5]{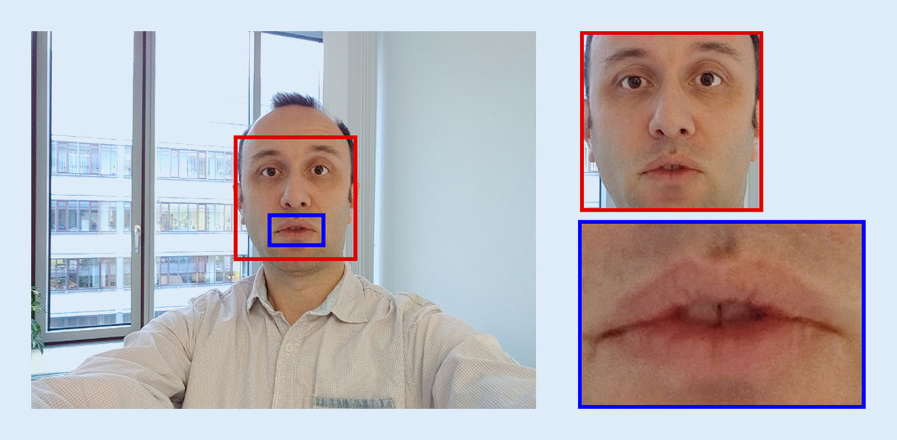}
\caption{Example of a video which was cropped to the face (red) and lips (blue).}
\label{Fig1:video_crop}
\end{figure}

The image signals from all videos were cropped to remove unnecessary information. All videos were uniformly either shortened or extended to a length of 28 frames by repeating the last frame and stored as 4-dimensional arrays (color channel, frame, height, width) normalized to [0,1]. In each frame, both the facial region and the mouth area were automatically identified, and each video was cropped accordingly. The detected facial regions were resized to 90 × 90 pixels, while the mouth regions were scaled to 150 × 100 pixels. Figure \ref{Fig1:video_crop} shows an example of the cropping.
Since different color spaces can lead to varying accuracies in image classification \cite{b13}, the videos were converted into the following color spaces: RGB, grayscale, HSV, LAB, XYZ, and YCbCr.

\subsection{Models and Training}

For training the neural network, the models 3D Convolutional Neural Network (Conv3D), Gated Recurrent Units (GRU) and GRUConv, a combination of the first two models, were used.

As the first layer of the Conv3D model, batch normalization \cite{b16} of the input data is performed so that the mean is 0 and the standard deviation is 1. This should improve the performance of the model and lead to a more stable training process. This is followed by a 3D CNN layer, MaxPooling and another batch normalization. All three layer types were repeated three times in this order. All three CNN and MaxPooling layers use a padding of (1, 2, 2). While all convolutional layers have a kernel size of (3, 5, 5), the MaxPooling layers use a kernel size of (1, 2, 2). The first 3D CNN is configured with a stride of (1, 2, 2), while the subsequent two use a stride of (1, 1, 1). The three CNN layers contain 8, 16, and 32 kernels respectively. After a final flattening and MLP, the model outputs a vector with 18 components. The softmax function was used to determine the probabilities for the word classes. Their maximum resulted in the word defined as output. The ReLU function was used as activation function for the 3D CNNs as well as the MLP. After each MaxPooling as well as at the MLP, a dropout \cite{b29} of 0.5 was applied. This means that random 50\% of the neurons in the layer were ignored during training.

\begin{figure}[h]
\centering
\includegraphics[scale=0.45]{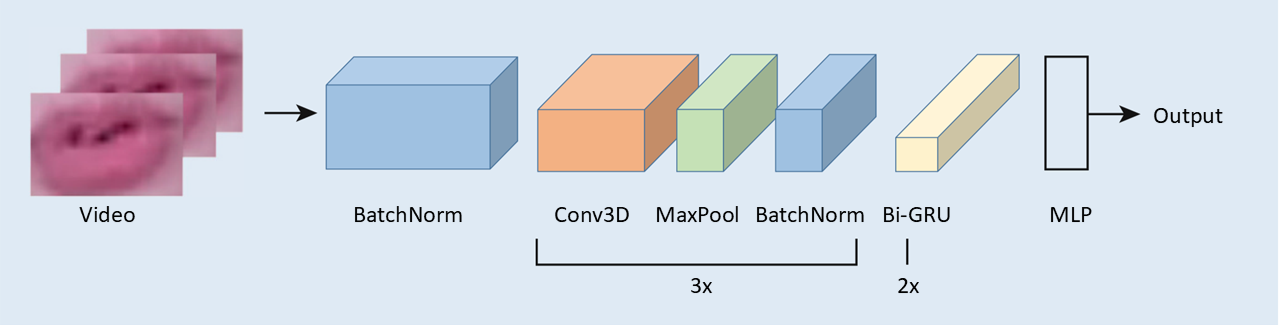}
\caption{Schema of the GRUConv model.}
\label{Fig2:model}
\end{figure}

Per training example and iteration in training, random neurons were always selected again. This reduced the risk of overfitting, where a model achieves high accuracies on the training set but achieves lower accuracies on unseen test data with progressing training. Figure \ref{Fig2:model} shows the sequence of layers used in the model.

In the GRU model, flattening was first performed per frame. With 3 channels and an image resolution of 150x100 pixels, this results in a vector with 45,000 elements per frame with a total of 28 frames. This data with a dimensionality of 28x45,000 was used as input for two bidirectional GRU layers. The outputs of the bidirectional GRUs were then combined per timestep to enable learning of sequences in reverse order as well. The output of the second GRU was finally fed into an MLP, which has the same configuration as the MLP of the Conv3D model.

In the GRUConv model, the two GRU layers of the GRU model were inserted between the layers of the Conv3D model and the MLP. While 3D CNNs extract spatial information, GRUs are well suited for temporal information.

\begin{figure}[h]
\centering
\includegraphics[scale=0.4]{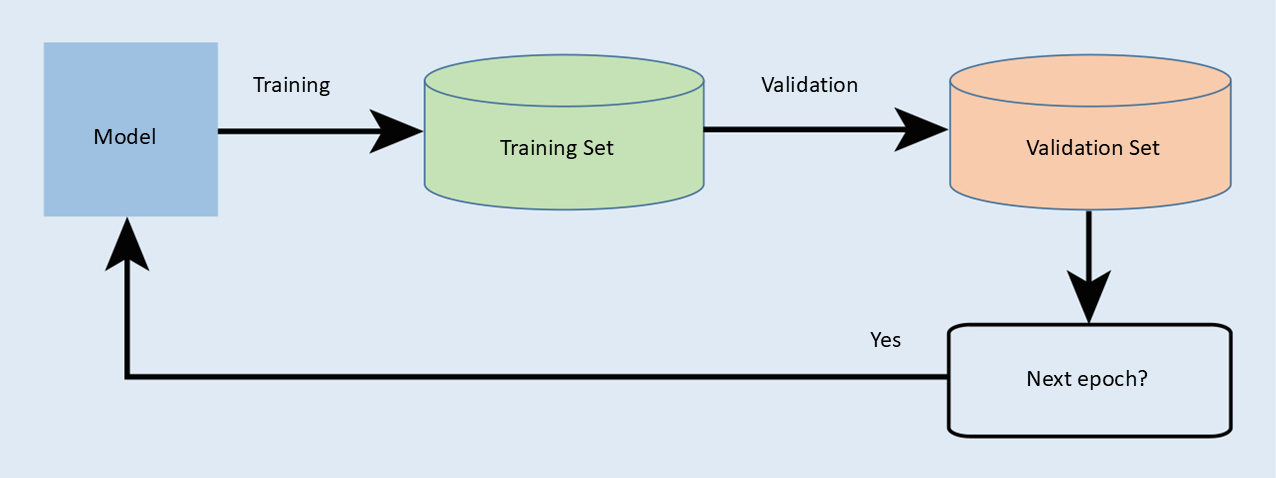}
\caption{Schema of the training strategy.}
\label{Fig3:training}
\end{figure}

During training of each of the models, multiple epochs were iterated. In each epoch, the model was trained with each video in the training set with a learning rate of $10^{-6}$, so that the learnable parameters of the neural network were updated. ADAM \cite{bDBLP:journals/corr/KingmaB14} was used as the optimization algorithm while cross entropy served as the loss function.
Subsequently, each video in the validation set was classified. The number of correctly classified videos divided by the total number of videos in the validation set yielded the accuracy of the epoch. The learned parameters in the epoch with the best validation accuracy were determined for each model and used for later evaluation. A batch size of 128 for the training set and 32 for the validation set was used.  Figure \ref{Fig3:training} schematically shows the described training strategy.

\subsection{Implementation}

The creation of the datasets, the processing pipeline, the models and the training process were implemented in the programming language Python. First, the YouTube links were manually collected for the dataset. Using youtube-dl \cite{b11}, the videos were subsequently downloaded. For speech recognition of the videos, the Python module of the Vosk API \cite{bvosk} was used, which includes an already existing language model for German. Since this in turn expects audio files as input, the open-source software FFmpeg \cite{b30} was used to extract the audio material (WAV files) from the videos (AVI files).

The Python module MoviePy \cite{b5} subsequently enabled the extraction of individual video sequences according to the recognized timestamps of the words. Afterwards, the videos were read for cropping the speakers using the OpenCV programming library \cite{b4} and were cropped to the faces through the Python module Face Recognition \cite{b10}. The extraction of the lip image sections themselves was implemented using the toolkits Dlib \cite{b17} and Imutils \cite{b25}. The division of the dataset into training, validation or test sets as well as the conversion of the videos into color spaces was performed using the Scikit-learn programming library \cite{b23}.

The AI models and their training were implemented using the PyTorch deep learning library \cite{b22}. As hardware, four Tesla V100-SXM2 graphics processors (Nvidia Corp., Santa Clara, USA) with 32GB memory each were used.

The lip-reading accuracy was determined within 5000 training epochs using the Conv3D model. For this, the two subsets of Dataset A (Table \ref{Tab:dataset}) were used for training and validation, comparing face cropping versus mouth region cropping in the RGB color space with respect to validation accuracy.

The image crop that achieved higher accuracy was then used in a subsequent step with the Conv3D model on the correspondingly cropped videos of Dataset A to compare the color spaces RGB, grayscale, HSV, LAB, XYZ, and YCbCr. The accuracy was again measured over 5000 training epochs.

The Conv3D, GRU, and GRUConv models were subsequently trained using Dataset A with the image crop and color space that had previously achieved the highest accuracies in each comparison, and were compared over 5000 training epochs based on validation accuracy.

As the final model for lip-reading, the neural network with the highest validation accuracy from the previous experiment was trained with Dataset B (Table \ref{Tab:dataset}). The cropping region and color space from the previous experiment were also adopted for the final model. The model was then applied to validation Dataset B in each of the 5000 epochs, so that the model parameters from the epoch with the highest validation accuracy were retained for the final model. The accuracy of the resulting final model was subsequently determined using the test dataset from Dataset B and Dataset C (Table \ref{Tab:dataset}) with unseen speakers. The lip-reading accuracy was measured as correct classification rate and presented as a confusion matrix.

\section{Results}

\begin{table}[]
\centering
\begin{tabular}{|l|llllll|}
\hline
\multirow{3}{*}{Word Class} & \multicolumn{6}{l|}{Frequency}                                                                                                                   \\ \cline{2-7} 
                            & \multicolumn{2}{l|}{Dataset A}                        & \multicolumn{2}{l|}{Datset B}                         & \multicolumn{2}{l|}{Dataset C}   \\ \cline{2-7} 
                            & \multicolumn{1}{l|}{Number} & \multicolumn{1}{l|}{\%} & \multicolumn{1}{l|}{Number} & \multicolumn{1}{l|}{\%} & \multicolumn{1}{l|}{Number} & \% \\ \hline
kommentare                  & \multicolumn{1}{l|}{118}    & \multicolumn{1}{l|}{3}  & \multicolumn{1}{l|}{387}    & \multicolumn{1}{l|}{1}  & \multicolumn{1}{l|}{72}     & 2  \\ \hline
fragen                      & \multicolumn{1}{l|}{171}    & \multicolumn{1}{l|}{5}  & \multicolumn{1}{l|}{918}    & \multicolumn{1}{l|}{3}  & \multicolumn{1}{l|}{154}    & 4  \\ \hline
prüfung                     & \multicolumn{1}{l|}{66}     & \multicolumn{1}{l|}{2}  & \multicolumn{1}{l|}{381}    & \multicolumn{1}{l|}{1}  & \multicolumn{1}{l|}{92}     & 2  \\ \hline
deutschland                 & \multicolumn{1}{l|}{56}     & \multicolumn{1}{l|}{2}  & \multicolumn{1}{l|}{422}    & \multicolumn{1}{l|}{1}  & \multicolumn{1}{l|}{138}    & 3  \\ \hline
können                      & \multicolumn{1}{l|}{325}    & \multicolumn{1}{l|}{9}  & \multicolumn{1}{l|}{2959}   & \multicolumn{1}{l|}{10} & \multicolumn{1}{l|}{215}    & 5  \\ \hline
sprechen                    & \multicolumn{1}{l|}{131}    & \multicolumn{1}{l|}{4}  & \multicolumn{1}{l|}{846}    & \multicolumn{1}{l|}{3}  & \multicolumn{1}{l|}{182}    & 5  \\ \hline
wirklich                    & \multicolumn{1}{l|}{159}    & \multicolumn{1}{l|}{4}  & \multicolumn{1}{l|}{3225}   & \multicolumn{1}{l|}{11} & \multicolumn{1}{l|}{174}    & 4  \\ \hline
eigentlich                  & \multicolumn{1}{l|}{194}    & \multicolumn{1}{l|}{5}  & \multicolumn{1}{l|}{1726}   & \multicolumn{1}{l|}{6}  & \multicolumn{1}{l|}{182}    & 5  \\ \hline
wissen                      & \multicolumn{1}{l|}{144}    & \multicolumn{1}{l|}{4}  & \multicolumn{1}{l|}{1110}   & \multicolumn{1}{l|}{4}  & \multicolumn{1}{l|}{65}     & 2  \\ \hline
natürlich                   & \multicolumn{1}{l|}{304}    & \multicolumn{1}{l|}{8}  & \multicolumn{1}{l|}{3212}   & \multicolumn{1}{l|}{10} & \multicolumn{1}{l|}{323}    & 8  \\ \hline
video                       & \multicolumn{1}{l|}{308}    & \multicolumn{1}{l|}{8}  & \multicolumn{1}{l|}{1855}   & \multicolumn{1}{l|}{6}  & \multicolumn{1}{l|}{264}    & 7  \\ \hline
bedeudet                    & \multicolumn{1}{l|}{118}    & \multicolumn{1}{l|}{3}  & \multicolumn{1}{l|}{790}    & \multicolumn{1}{l|}{3}  & \multicolumn{1}{l|}{295}    & 7  \\ \hline
beispiel                    & \multicolumn{1}{l|}{585}    & \multicolumn{1}{l|}{16} & \multicolumn{1}{l|}{3363}   & \multicolumn{1}{l|}{11} & \multicolumn{1}{l|}{960}    & 24 \\ \hline
schreiben                   & \multicolumn{1}{l|}{106}    & \multicolumn{1}{l|}{3}  & \multicolumn{1}{l|}{932}    & \multicolumn{1}{l|}{3}  & \multicolumn{1}{l|}{158}    & 4  \\ \hline
menschen                    & \multicolumn{1}{l|}{274}    & \multicolumn{1}{l|}{7}  & \multicolumn{1}{l|}{2385}   & \multicolumn{1}{l|}{8}  & \multicolumn{1}{l|}{62}     & 2  \\ \hline
einfach                     & \multicolumn{1}{l|}{355}    & \multicolumn{1}{l|}{10} & \multicolumn{1}{l|}{4435}   & \multicolumn{1}{l|}{14} & \multicolumn{1}{l|}{360}    & 9  \\ \hline
wichtig                     & \multicolumn{1}{l|}{146}    & \multicolumn{1}{l|}{4}  & \multicolumn{1}{l|}{1501}   & \multicolumn{1}{l|}{5}  & \multicolumn{1}{l|}{118}    & 3  \\ \hline
wörter                      & \multicolumn{1}{l|}{167}    & \multicolumn{1}{l|}{4}  & \multicolumn{1}{l|}{237}    & \multicolumn{1}{l|}{1}  & \multicolumn{1}{l|}{136}    & 3  \\ \hline
\end{tabular}
\caption{Frequency of videos per word class.}
\label{Tab:words}
\end{table}

\begin{figure}[h]
\centering
\includegraphics[scale=0.48]{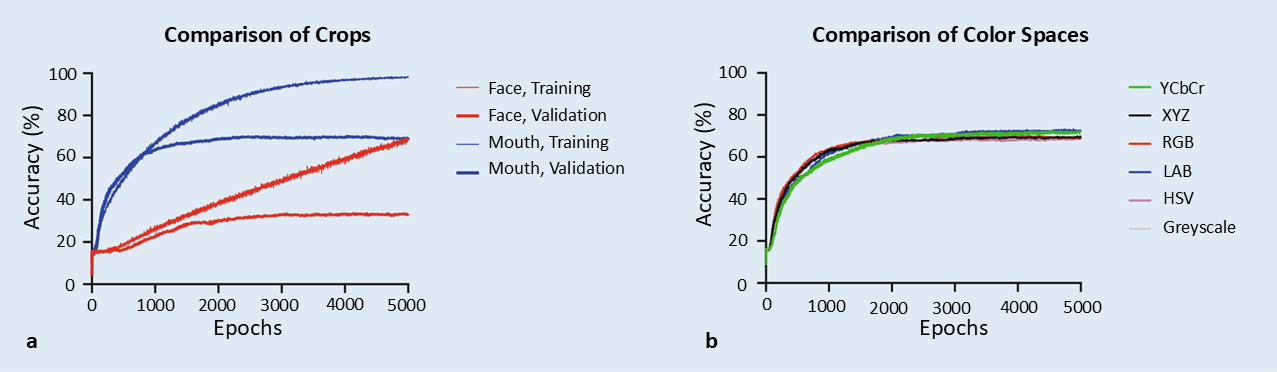}
\caption{Word recognition accuracy (correct classification rate) versus number of training epochs comparing (a) image crops and (b) color spaces.}
\label{Fig4:plot}
\end{figure}

Table \ref{Tab:words} summarizes the video frequencies per word class. Figure \ref{Fig4:plot}a shows the performance of the Conv3D model for the training and validation set of dataset A over 5000 consecutive epochs, comparing different image crops. When applied to the training set with face cropping, a continuous linear accuracy improvement of the word recognition up to 69.69\% after 5000 epochs was observed. With lip cropping, the model reached saturation at 98.62\% after 5000 epochs. For face-cropped videos, the model achieved 33.82\% validation accuracy after 5000 epochs. With lip cropping, a maximum accuracy of 70.29\% was reached after just 2454 epochs. Both cropping methods lead to overfitting, evidenced by saturation or decrease in validation accuracy despite increasing training accuracy.

Figure \ref{Fig4:plot}b displays the Conv3D model's performance on dataset A over 5000 epochs across different color spaces, showing nonlinear growth with saturation around 2000 epochs. Maximum validation accuracies were: 70.29\% (2454 epochs, RGB), 70.56\% (4970 epochs, grayscale), 69.23\% (4848 epochs, HSV), 73.47\% (4684 epochs, LAB), 72.15\% (4510 epochs, YCbCr), and 69.89\% (3928 epochs, XYZ).

When comparing models using lip crops and LAB color space, peak validation accuracies were: 73.47\% (Conv3D, 4684 epochs), 60.34\% (GRU, 2466 epochs), and 77.59\%(GRUConv, 4297 epochs). Thus, the GRUConv model achieved highest accuracy.

\begin{figure}[h]
\centering
\includegraphics[scale=0.48]{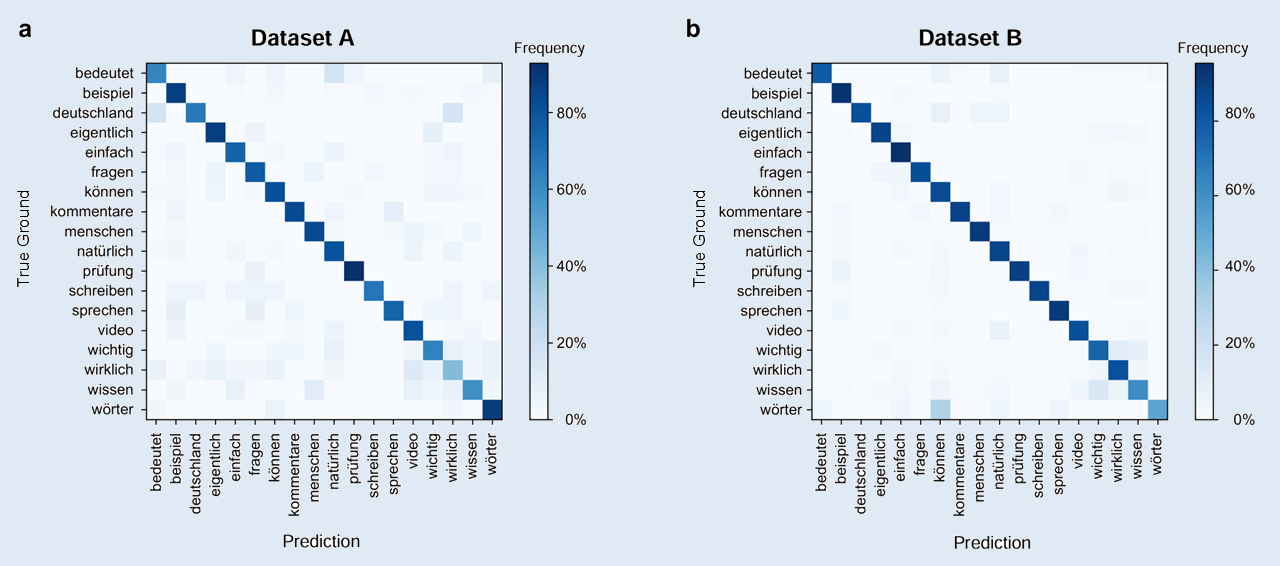}
\caption{Confusion matrices of predicted versus true word classes for (a) Dataset A and (b) Dataset B using GRUConv. Entries show prediction frequencies for each true class (y-axis) across predicted classes (x-axis). Diagonal entries indicate correct predictions; off-diagonal entries show misclassifications.}
\label{Fig5:confusion}
\end{figure}

Applied to training and validation dataset B, the GRUConv model reached maximum accuracy of 86.73\% after 1868 epochs, showing improved performance over dataset A. On dataset B's test set, it achieved 87.3\% accuracy. Figure \ref{Fig5:confusion} shows confusion matrices comparing predicted versus actual word classes for datasets A and B using GRUConv. Qualitative analysis reveals minimal class confusion, indicating high model sensitivity and specificity. When applied to dataset C with unseen speakers, the GRUConv model achieved 62.61\% accuracy.

\section{Discussion}

This work describes a deep learning algorithm that classifies words based on mouth movements with high accuracy in a defined dataset of video recordings.

A crucial processing step is the selection of the image crop. Here, cropping to the mouth region resulted in significantly higher correct classification rates compared to face cropping. One reason might be that the facial area contains too much irrelevant information that complicates classification. When using the entire face as the image crop, a model with different hyperparameters, architecture, or more neurons would be necessary due to the more complex function to approximate. The facial image resolution of 90×90 pixels could also contribute to the poorer performance, as scaling might cause loss of relevant information. Higher resolution would however massively increase computational requirements. The use of lip regions has become standard in comparable English-language studies, such as in LipNet \cite{b2} and Google DeepMind \cite{b27}. On the other hand, using face-cropped videos, particularly with landmark detectors \cite{b15}, can also achieve good performance, sometimes even surpassing state-of-the-art models using lip crops \cite{b8}.

Regarding color spaces, minimal differences in classification accuracy were observed. This is expected since mouth region images inherently have limited color diversity. Gowda \& Yuan \cite{b13} show that models perform similarly across color spaces even with diverse object datasets. There too, the LAB color space achieved highest accuracy, with less than 2\% difference to the second-best space. In this work, we cannot exclude that certain word classes might be classified more accurately in specific color spaces. This could explain why grayscale images, containing presumably less information, outperformed RGB. Different network architectures might achieve higher accuracy with certain color spaces. Additionally, random weight initialization could have influenced the results.

Among the tested deep learning models, the GRUConv model combining CNN and GRU achieved highest classification accuracy for lip-reading. Since the underlying GRU and CNN layer types remained unchanged, this suggests synergistic effects from their combination.

The final model on dataset A demonstrates that German lip-reading using neural networks is feasible. The 77.59\% accuracy is comparable to results achieved for other languages \cite{b14}.

A major challenge in lip-reading is word confusions. In our test set, the words "wichtig" (important), "wirklich" (really), and "wissen" (know) had lowest recognition rates and were most frequently confused. This stems from both similar mouth movements and imbalanced class distribution. Conversely, classes with fewer videos like "prüfung" (exam) and "deutschland" (Germany) showed relatively high accuracy due to their distinctive articulations. By contrast, the word classes 'wichtig' (important), 'wirklich' (really), and 'wissen' (know) have relatively few training videos available, yet exhibit similar lip movements. The larger dataset B confirmed that more training examples improve differentiation, though new confusions emerged (e.g., "wörter" [words] with "können" [can]), likely due to articulatory similarities.

Overall, as with machine learning generally, training data quantity proved to be crucial for lip-reading. Our results show dataset size impacts accuracy more than model choice.

The final model demonstrated good generalization for known speakers and was finally evaluated on dataset C. There, the decrreased accuracy of 62.61\% was to be expected for unseen speakers, but it is still high. Thus, a certain speaker-independent generalizability of the model can be assumed.

Potential errors in dataset creation (speech recognition or automatic cropping) might have influenced results, though manual verification showed high precision in test sets. An advantage of the used videos is that they represent natural daily speech from in-the-wild sources, increasing the relevance of this work for real life communication.

The final model's accuracy (87.3\% known speakers, 62.61\% unknown) demonstrates viable German lip-reading via deep learning. The accuracy of the model was successfully improved by expanding the training dataset. Future work could classify more than 18 word classes, apply more complex architectures with more training data, optimize hyperparameters, and implement Connectionist Temporal Classification for variable-length video input. Potential applications include automated subtitles for hearing-impaired users, silent dictation in noisy environments, and speech recognition system augmentation.

\section{Conclusion}

In this work, a neural network for lipreading in German was successfully implemented. A model was trained whose accuracies validated the suitability of the datasets as well as the utilized methods. The data sources and the used speech and mouth recognition proved to be effective.

\section*{Acknowledgement}

The authors thank Prof. Dr. Korinna Bade and Prof. Dr. Mike Scherfner (Hochschule Anhalt, Köthen, Germany) for providing computing resources and consultation. Furthermore, we are grateful to the Gymnasium Philanthropinum (Dessau-Roßlau, Germany), particularly Mr. Ron Seidel, and Stiftung Jugend forscht e.V. (Hamburg, Germany) for establishing the conditions for this project.

\bibliographystyle{unsrt}  
\bibliography{references}

\end{document}